\documentclass{article}
\usepackage{spconf,amsmath,epsfig}
\pdfoutput=1
\let\OLDthebibliography\thebibliography
\renewcommand\thebibliography[1]{
  \OLDthebibliography{#1}
  \setlength{\parskip}{0pt}
  \setlength{\itemsep}{0pt plus 0.3ex}
}

\usepackage{amsfonts}
\usepackage{amssymb}
\usepackage{amsthm,amsmath}
\usepackage{mathrsfs}
\usepackage{indentfirst}
\usepackage{multirow}
\usepackage{algorithm}  
\usepackage{algorithmic}
\usepackage[utf8]{inputenc}
\usepackage{soul}
\usepackage{booktabs}

\usepackage{amsmath}
\usepackage{amsthm}
\usepackage{amssymb}

\usepackage{color}
\usepackage{xcolor}
\usepackage{subfigure}
\usepackage{bm}

\usepackage{colortbl}

\definecolor{mygray}{gray}{.8}

\definecolor{mypink}{rgb}{.99,.91,.95}

\definecolor{mycyan}{cmyk}{.3,0,0,0}

\def\etal{{\em et al.\/}\, }

\def\ie{\textit{i.e.}}
\usepackage[colorlinks,linkcolor=black,citecolor=black]{hyperref}

\usepackage{algorithm}
\usepackage{algorithmic}

\usepackage{multirow}

\def\mb{\mathbf}
\def\mbb{\mathbb}
\def\mc{\mathcal}

\usepackage{newfloat}
\usepackage{listings}
\floatstyle{ruled}
\newfloat{listing}{tb}{lst}{}
\floatname{listing}{Listing}

\begin{document}\sloppy

\def\x{{\mathbf x}}
\def\L{{\cal L}}


\title{AGCN: Augmented Graph Convolutional Network\\for Lifelong Multi-label Image Recognition}

%

\name{Kaile Du$^{1*}$, Fan Lyu$^{1,2*}$, Fuyuan Hu$^{1\dagger}$, Linyan Li$^{3\dagger}$, Wei Feng$^{2}$, Fenglei Xu$^{1}$, Qiming Fu$^{1}$}
\address{$^{1}$Suzhou University of Science and Technology, China\\
$^{2}$Tianjin University, China\\
$^{3}$Suzhou Institute of Trade \& Commerce, China\\}
\maketitle

\begin{abstract}
	The Lifelong Multi-Label (LML) image recognition builds an online class-incremental classifier in a sequential multi-label image recognition data stream.
  	The key challenges of LML image recognition are the construction of label relationships on \textit{Partial Labels} of training data and the \textit{Catastrophic Forgetting} on old classes,  resulting in poor generalization.	
  	To solve the problems, the study proposes an Augmented Graph Convolutional Network (AGCN) model that can construct the label relationships across the sequential recognition tasks and sustain the catastrophic forgetting.
  	First, we build an Augmented Correlation Matrix (ACM) across all seen classes, where the intra-task relationships derive from the hard label statistics while the inter-task relationships leverage both hard and soft labels from data and a constructed expert network.
	Then, based on the ACM, the proposed AGCN captures label dependencies with dynamic augmented structure and yields effective class representations.
	Last, to suppress the forgetting of label dependencies across old tasks, we propose a relationship-preserving loss as a constraint to the construction of label relationships.
	The proposed method is evaluated using two multi-label image benchmarks and the experimental results show that the proposed method is effective for LML image recognition and can build convincing correlation across tasks even if the labels of previous tasks are missing. Our code is available at https://github.com/Kaile-Du/AGCN.
\end{abstract}

\begin{keywords}
Lifelong Multi-Label Image Recognition, Graph Convolutional Network, Augmented Correlation Matrix, Relationship-preserving Loss
\end{keywords}

\renewcommand{\thefootnote}{\fnsymbol{footnote}}
\footnotetext[1]{Co-first author.}
\footnotetext[2]{Corresponding author.}

\section{Introduction}

\begin{figure}[t]
	\centering
	\includegraphics[width=\linewidth]{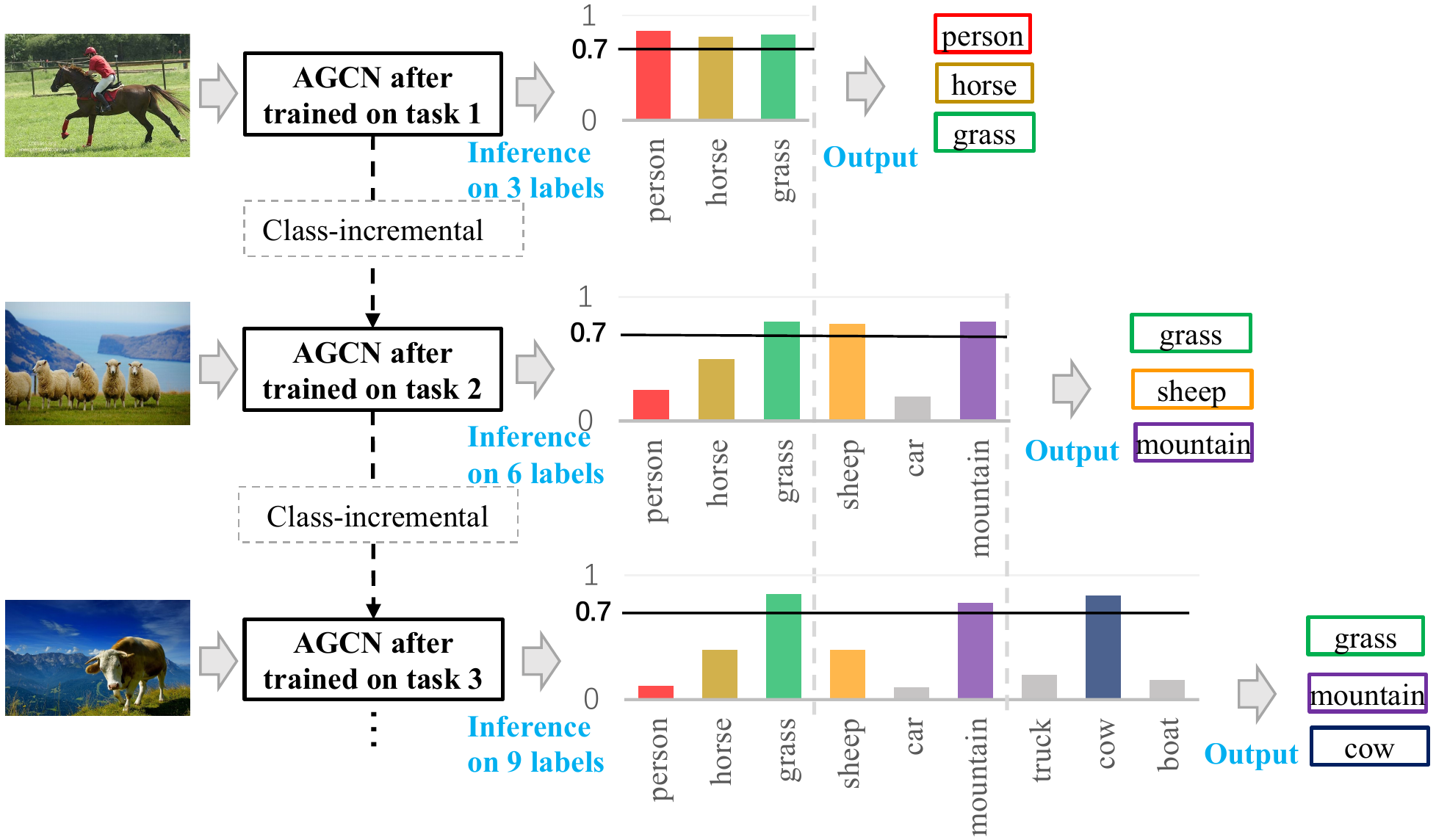}
	\caption{The inference of LML image recognition. Given multi-label images for testing, the model can recognize more labels by learning more incremental classes (label with the probability that greater than the threshold 0.7 will be output).
	}
	\label{fig:lml}
	\vspace{-15px}
\end{figure}

Class-incremental image recognition task~\cite{rebuffi2017icarl} constructs a unified evolvable classifier, which online learns new classes from a sequential image data stream and achieves multi-label classification for the seen classes.
However, most existing lifelong learning studies~\cite{kirkpatrick2017overcoming,li2017learning,bang2021rainbow,lyu2020multi,kim2020imbalanced,fernando2017pathnet,mallya2018packnet} only consider the input images are single-labelled (Lifelong Single-Label, LSL), which introduces significant limitations in practical applications such as the movie categorization and the scene classification.
This paper studies how to sequentially learn classes from new tasks for the Lifelong Multi-Label (LML) image recognition. 
As shown in Fig.~\ref{fig:lml}, given testing images, the model can continuously recognize more multiple labels with new classes learned.

For privacy, storage and efficient-computation reasons, the training data in lifelong learning for the old tasks are often unavailable when new tasks arrive, and the new task data only has labels of itself.
Thus, the catastrophic forgetting~\cite{mccloskey1989catastrophic}, \ie, the training on new tasks may lead to the old knowledge overlapped by the new knowledge, is the main challenge of LSL image recognition.
However, it is stated that LML image recognition is challenging due to not just catastrophic forgetting, but
\textit{Partial labels} for the current tasks, which means the training image may contain possible labels of past and future tasks.
To the best of our knowledge, there exist few lifelong learning algorithms designed specifically for LML image recognition against this challenge.


In this paper, inspired by the recent research on label relationships in multi-label learning~\cite{chen2019multi}, we consider building the label relationships across tasks.
\textit{However, because of the partial label problem, it is difficult to construct the class relationships by using statistics directly.}
To deal with the partial label problem, we propose an AGCN, a novel solution to LML image recognition. 
Our method saves no past data, and every data point passes only once.
First, an auto-updated expert network is designed to generate predictions of the old tasks, these predictions as soft labels are used to represent the old classes for the old tasks and to construct the ACM.  Then, the AGCN receives the dynamic ACM and correlates the label spaces of both the old and new tasks, which continually supports the multi-label prediction. 
Moreover, to further mitigate the forgetting on both seen classes and class relationships, a distillation loss and a relationship-preserving loss function are designed for both the class-level forgetting and relationship-level forgetting, respectively.
We construct two multi-label image datasets Split-COCO and Split-WIDE based on MS-COCO and NUS-WIDE, respectively. 
The results on Split-COCO and Split-WIDE show that our AGCN achieves state-of-art performances in LML image recognition.

Our technical contributions are two-fold:

1) We propose an AGCN model for LML image recognition to analyze the dynamic ACM to construct label relationships in the data stream.

2) We propose a relationship-preserving loss to mitigate the catastrophic forgetting phenomenon that happens on the label relationship level.

\section{Related Work}



\textbf{Class-incremental lifelong learning}.
To solve the catastrophic forgetting problem, the state-of-art methods for class-incremental lifelong learning can be categorized into three main types.
First, the regularization-based methods~\cite{kirkpatrick2017overcoming,hinton2015distilling}, which are based on regularizing the parameters corresponding to the old tasks and penalizing the feature drift on the old tasks.
For instance, Kirkpatrick \etal\cite{kirkpatrick2017overcoming} limited changes to parameters based on their significance to the previous tasks using Fisher information;
\cite{li2017learning} leveraged the knowledge distillation combined with standard cross-entropy loss~\cite{hinton2015distilling} to avoid forgetting by storing the previous parameters.
Second, the rehearsal-based methods~\cite{bang2021rainbow,lyu2020multi,rebuffi2017icarl,rolnick2019experience,chaudhry2018efficient,chaudhry2019tiny}, which sample a subset of data from the previous tasks as the memory. 
For example, in ER~\cite{rolnick2019experience}, this memory was retrained as the extended training dataset during the current training; 
AGEM~\cite{chaudhry2018efficient} reset the training gradient by combining the gradient on the memory and training data; RM~\cite{bang2021rainbow} is a replay method in the blurry setting;
Third, the parameter isolation based methods~\cite{fernando2017pathnet,mallya2018packnet,serra2018overcoming}, which generated task-specific parameter expansion or sub-branch. 
Though the existing methods have achieved significant successes in LSL, they are hardly be used in LML image recognition directly.


\noindent
\textbf{Multi-label image classification}.
Some methods \cite{lyu2018coarse,lyu2019multi} focused on global/local attention for multi-label learning. Recent advances are mainly by constructing label relationships.
Some methods \cite{wang2016cnn,jin2016annotation,lyu2019attend} used recurrent neural network multi-label recognition under a restrictive assumption that the label relationships are in order, which limited the complex relationships in label space. 
Furthermore, some methods \cite{chen2019multi,chen2020knowledge} built the label relationships using the graph structure and used graph convolutional network (GCN) to enhance the representation.
The common limit of these methods is that they can only construct the intra-task correlation matrix using the training data from the current task, and fail to capture the inter-task label dependencies in a lifelong data stream. 
Kim \etal\cite{kim2020imbalanced} proposed to extend the ER~\cite{rolnick2019experience} algorithm using a different sampling strategy for rehearsal on multi-label datasets. However, the label dependencies were ignored in this work \cite{kim2020imbalanced}. 
In contrast, we propose to model the label relationships and consider mitigating the relationship-level forgetting in LML image recognition.

\section{Methodology}

\begin{figure*}[t]
	\centering
	\includegraphics[width=\linewidth]{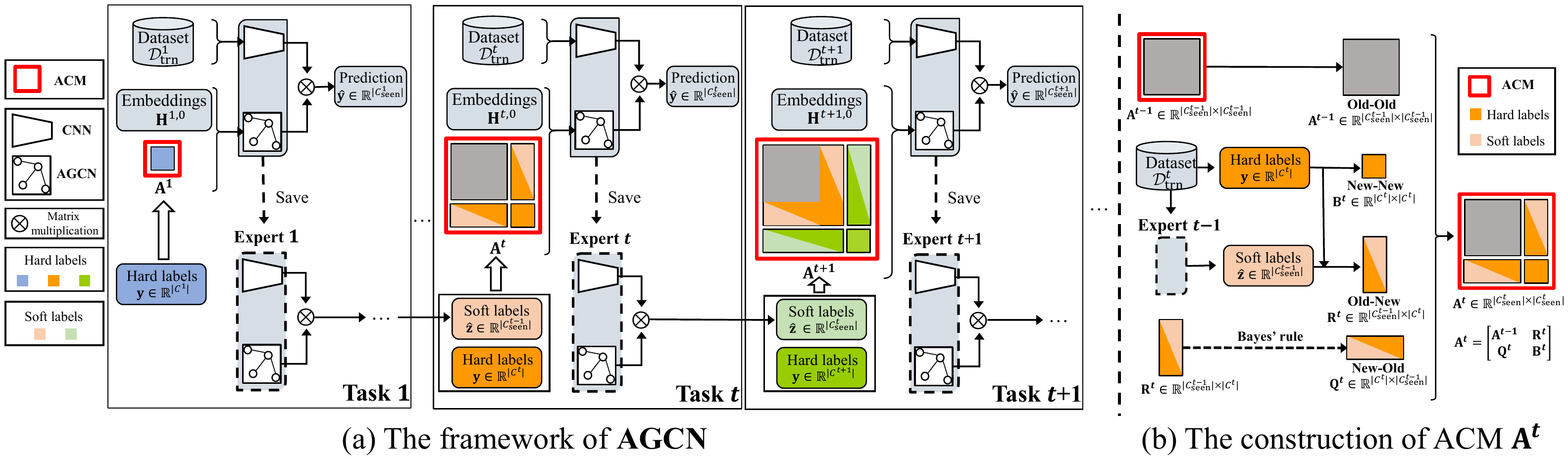}
	\vspace{-20px}
	\caption{
		The framework of AGCN (a) and the construction of ACM (b). 
		In (a), the training data $\mc{D}^t_\text{trn}$  for task $t$  is fed into CNN block, and the graph node embeddings $\mb{H}^{t,0}$ and the $\mb{A}^t$ are input to the AGCN block. 
		After each task has been trained, we save expert blocks to provide soft labels $\mb{\hat{z}}$ in the next task training. 
		In (b), the ACM is constructed using soft and hard labels in four blocks, which represents Old-Old, New-New, Old-New and New-Old blocks.
	}
	\label{fig:framework}
	\vspace{-10px}
\end{figure*}


\subsection{Lifelong multi-label learning}

We provide the definition of LML image recognition.
In this study, each data is trained only once in the form of a data stream. 
Given $T$ recognition tasks with respect to train datasets $\{\mc{D}^{1}_\text{trn},\cdots,\mc{D}^{T}_\text{trn}\}$ and test datasets $\{\mc{D}^{1}_\text{tst},\cdots,\mc{D}^{T}_\text{tst}\}$. 
In LML image recognition, the total class numbers increase gradually with the sequential tasks. 
For the $t$-th task, we have new and task-specific classes to be trained namely $\mc{C}^t$. 
\textit{The goal is to build a multi-label classifier to discriminate increasing number of classes.}
We denote $\mc{C}_\text{seen}^t=\bigcup_{n=1}^{t}\mc{C}^n$ as seen classes at task $t$, where $\mc{C}_\text{seen}^t$ contains old class set $\mc{C}_\text{seen}^{t-1}$ and new class set $\mc{C}^t$, that is, $\mc{C}_\text{seen}^t=\mc{C}_\text{seen}^{t-1}\cup\mc{C}^t$.
Note that during the testing phase, the ground truth labels for LML image recognition contain all the old classes $\mc{C}_\text{seen}^t$.
\subsection{Overview of the proposed method}
In traditional multi-label learning, label relationships are verified effective to improve the recognition~\cite{zhu2017multi,zhang2018binary}, or about generation~\cite{zhao2021each}.
However, it is still challenging to construct convincing label relationships in LML image recognition because of the \textit{partial labels} of every task, the old classes are unavailable, which results in the difficulty of constructing the inter-task label relationships.
Moreover, the forgetting happens not only on the class level but also the relationship level, which may damage the performance. 
At each step of the training in an online fashion, we propose an AGCN to construct and update the intra- and inter-task label relationships and we also propose a relationship-preserving loss to mitigate the relationship-level forgetting.

As shown in Fig.~\ref{fig:framework} (a), the proposed method consists of two main components: 
1) \textbf{Augmented Correlation Matrix (ACM)}  provides the label relationships among all seen classes and is augmented to capture the intra- and inter-task label independences.
2) \textbf{Augmented Graph Convolutional Network (AGCN)} provides  structural label representations for label relationships. 
Together with an CNN feature extractor, the multiple labels for an image $\mb{x}$ will be predicted by
\begin{equation}\label{eq:predict}
	\mb{\hat{y}} = \sigma\left({\text{AGCN}(\mb{A}^t, \mb{H}^{t,0})}\otimes \text{CNN}\left(\mb{x}\right)\right),
\end{equation}
where $\mb{A}^t$ denotes the ACM and $\mb{H}^{t,0}$ is the initialized graph node.
$\otimes$ denotes the matrix multiplication and $\sigma(\cdot)$ represents the sigmoid function to classify.
Suppose $D$ represents the image feature dimensionality, because $\text{AGCN}(\mb{A}^t, \mb{H}^{t,0})\in\mbb{R}^{|\mc{C}_\text{seen}^t| \times D}$ and $\text{CNN}(\mb{x})\in\mbb{R}^{ D}$, we have the prediction $\mb{\hat{y}} = [\mb{\hat{y}}_\text{old}~\mb{\hat{y}}_\text{new}]$, where $\mb{\hat{y}}_\text{old}\in\mbb{R}^{|\mc{C}_\text{seen}^{t-1}|}$ for old classes and $\mb{\hat{y}}_\text{new}\in\mbb{R}^{|\mc{C}^t|}$ for new classes for $t>1$. By binarizing the ground truth to hard labels $\mb{y} = [y_1, \cdots, y_{|\mc{C}^t|}]^\top, y_i\in\{0,1\}$, we train the current task for classifying using the Cross Entropy loss:
\begin{equation}\label{eq:current_loss}	
		\ell_\text{cls}(\mb{y},\mb{\hat{y}}_\text{new})=-
		\sum_{i=1}^{|\mc{C}^t|}\Big[y_i\log\left({{\hat{y}}}_{i}\right)+\left(1-{{y}}_{i}\right)\log\left(1-{{\hat{y}}}_{i}\right)\Big].
\end{equation}
To mitigate the class-level catastrophic forgetting, inspired by the distillation-based lifelong learning method~\cite{li2017learning}, we construct auto-updated expert networks consisting of CNN$_\text{xpt}$ and AGCN$_\text{xpt}$.
The expert parameters are fixed after each task has been trained and auto-update along with new task learning.
Based on the expert, we construct the distillation loss as
\begin{equation}\label{eq:distillation}
		\ell_\text{dst}({\mb{\hat{z}}},\mb{\hat{y}}_\text{old})=-\sum_{i=1}^{|\mc{C}_\text{seen}^{t-1}|}\left[{{\hat{z}_i}}\log\left({\hat{y}}_{i}\right)+\left(1-{{\hat{z}}}_{i}\right)\log\left(1-{\hat{y}}_{i}\right)\right],
\end{equation}
where ${\mb{\hat{z}}} = \sigma\left(\text{AGCN}_\text{xpt}(\mb{A}^{t-1}, \mb{H}^{t-1,0}) \otimes \text{CNN}_\text{xpt}\left(\mb{x}\right)\right)$ can be treated as the \textit{soft labels} to represent the prediction on old classes.
The $i$-th element $\hat{z}_i$ of ${\mb{\hat{z}}}$ represent the probability  that the image $\mb{x}$ contains the class.
Then, the major problems are how to construct the ACM $\mb{A}$ (Sec~\ref{sec:acm}) and implement the training of AGCN (Sec~\ref{sec:agcn}).

\subsection{Augmented Correlation Matrix}

\label{sec:acm}

Most existing multi-label learning algorithms~\cite{chen2019multi} rely on constructing the inferring label correlation matrix $\mb{A}$ by the hard label statistics among the class set $\mc{C}$: $\mb{A}_{ij}=P(\mc{C}_i|\mc{C}_j)|_{i \neq j}$.
As shown in Fig.~\ref{fig:framework} (b), we construct ACM $\mb{A}^t$ for task $t>1$ in an online fashion to simulate the statistic value denoted as
\begin{equation}\label{eq:acm}
	{\mb{A}}^{t}=\begin{bmatrix} {\mb{A}}^{t-1} & \mb{R}^t \\ \mb{Q}^t & \mb{B}^t \end{bmatrix}=\begin{bmatrix} \text{Old-Old} & \text{Old-New} \\ \text{New-Old} & \text{New-New} \end{bmatrix},
\end{equation}
in which we take four block matrices including $\mb{A}^{t-1}$ and $\mb{B}^t$, $\mb{R}^t$ and $\mb{Q}^t$ to represent intra- and inter-task label relationships between old and old classes, new and new classes, old and new classes as well as new and old classes respectively.
For the first task, $\mb{A}^1=\mb{B}^1$.
For $t>1$, $\mb{A}^t\in\mbb{R}^{|\mc{C}_\text{seen}^t|\times|\mc{C}_\text{seen}^t|}$.
It is worth noting that the block $\mb{A}^{t-1}$ (Old-Old) can be derived directly from the old task, so we will focus on how to compute the other three blocks in the ACM.


\noindent
\textbf{New-New block} ($\mb{B}^t\in\mbb{R}^{|\mc{C}^t|\times|\mc{C}^t|}$).
This block computes the intra-task label relationships among the new classes, and the conditional probability in $\mb{B}^t$ can be calculated using the hard label statistics from the training dataset similar to the common multi-label learning: 
\begin{equation}\label{eq:hard-hard}
	\mb{B}^t_{ij} = P(\mc{C}_{i}\in\mc{C}^t|\mc{C}_{j}\in\mc{C}^t) = \frac{N_{ij}}{N_{j}},
\end{equation}
where $N_{ij}$ is the number of examples with both class $\mc{C}_{i}$ and $\mc{C}_{j}$, $N_{j}$ is the number of examples with class $\mc{C}_{j}$. 
Due to the online data stream, $N_{ij}$ and $N_{j}$ are accumulated and updated at each step of the training process.



\noindent
\textbf{Old-New block} ($\mb{R}^t\in\mbb{R}^{|\mc{C}_\text{seen}^{t-1}|\times|\mc{C}^t|}$). 
{Given an image $\mb{x}$, for the old classes, ${\hat{z}}_{i}$ (predicted probability) generated by the expert  can be considered as the soft label for the $i$-th class (see Eq.~\eqref{eq:distillation}). 
Thus, the product ${\hat{z}}_{i}{{{y}}_{j}}$ can be regarded as an alternative of the cooccurrences of $\mc{C}_{i}$ and $\mc{C}_{j}$, \ie, $N_{ij}$. 
$\sum_{\mb{x}}{\hat{z}}_{i}{{{y}}_{j}}$ means the online mini-batch accumulation.
Thus, we have
\begin{equation}\label{eq:soft-hard}
	\begin{aligned}		
	\mb{R}^t_{ij} 	&= P(\mc{C}_{i}\in{\mc{C}_\text{seen}^{t-1}}|\mc{C}_{j}\in\mc{C}^t) = \frac{\sum_{\mb{x}}{\hat{z}}_{i}{{{y}}_{j}}}{N_j}.
	\end{aligned}
\end{equation}

\noindent
\textbf{New-Old block} ($\mb{Q}^t\in\mbb{R}^{|\mc{C}^t|\times|\mc{C}_\text{seen}^{t-1}|}$). 
Based on  Bayes' rule, we can obtain this block by
\begin{equation}\label{eq:hard-soft}
	\begin{aligned}
		\mb{Q}^t_{ji} &= P(\mc{C}_{j}\in\mc{C}^t|\mc{C}_{i}\in{\mc{C}_\text{seen}^{t-1}}) = \frac{P(\mc{C}_i|\mc{C}_j)P(\mc{C}_j)}{P(\mc{C}_i)}=\frac{\mb{R}^t_{ij}N_j}{\sum_{\mb{x}}{\hat{z}}_{i}}.
	\end{aligned}
\end{equation}
Finally, we online construct an ACM using the soft label statistics from the auto-updated expert network and the hard label statistics from the training data. 

\subsection{Augmented Graph Convolutional Network}

\label{sec:agcn}

ACM is auto-updated dependencies among all seen classes in the LML image recognition system.
With the established ACM, we can leverage Graph Convolutional Network (GCN) to assist the prediction of CNN as Eq.~\eqref{eq:predict}.
{We propose an Augmented Graph Convolutional Network (AGCN) to manage the augmented fully-connected graph.} 
AGCN is a two-layer stacked graph model, which is similar to ML-GCN~\cite{chen2019multi}, and more details can be found in the supplementary material.
Based on the ACM $\mb{A}^t$, AGCN can capture class-incremental dependencies in an online way.
Let the graph node be initialized by the Glove embedding~\cite{pennington2014glove} namely $\mb{H}^{t,0}\in\mbb{R}^{|\mc{C}_\text{seen}^t|\times d}$ where $d$ represents the embedding dimensionality.
The graph presentation $\mb{H}^t \in\mbb{R}^{|\mc{C}_\text{seen}^t| \times D}$ in task $t$ is mapped by:
\begin{equation}\label{eq:agcn}
		\mb{H}^t=\text{AGCN}(\mb{A}^t, \mb{H}^{t,0}).
\end{equation}
To mitigate the relationship-level forgetting across tasks, we constantly preserve the established relationships in the sequential tasks. 
In contrast to saving raw data, the graph node embedding is irrelevant to the label co-occurrence and can be stored as a teacher to avoid the forgetting of label relationships.
Suppose the learned embedding after task $t$ is stored as $\mb{G}^{t}=\text{AGCN}_\text{xpt}(\mb{A}^{t}, \mb{H}^{t,0})$, $t>1$.
We propose a relationship-preserving loss as a constraint to the class relationships:
\begin{equation}\label{eq:graph_distillation}
		\ell_\text{gph}({\mb{G}}^{t-1},\mb{H}^{t})=\sum^{|\mc{C}_\text{seen}^{t-1}|}_i \left\Vert{\mb{G}}^{t-1}_i-\mb{H}^{t}_i\right\Vert^2.
\end{equation}
By minimizing $\ell_\text{gph}$ with the partial constraint of old node embedding, the changes of $\text{AGCN}$ parameters are limited.
Thus, the forgetting of the established label relationships are alleviated with the progress of LML image recognition. 
The final loss for the model training is defined as
\begin{equation}\label{eq:final_loss}
	\ell=\lambda_1\ell_\text{cls}(\mb{y},\mb{\hat{y}}_\text{new})+\lambda_2\ell_\text{dst}(\mb{\hat{z}},\mb{\hat{y}}_\text{old})+ \lambda_3\ell_\text{gph}({\mb{G}}^{t-1},\mb{H}^{t}),
\end{equation}
where $\ell_\text{cls}$ is the classification loss, $\ell_\text{dst}$ is used to mitigate the class-level forgetting and $\ell_\text{gph}$ is used to reduce the  relationship-level forgetting. $\lambda_1$, $\lambda_2$ and $\lambda_3$ are the loss weights for $\ell_\text{cls}$, $\ell_\text{dst}$ and $\ell_\text{gph}$, respectively. Making $\lambda_1$ larger will favor the new task performance, making $\lambda_2$ larger will favor the old task performance. Extensive  ablation studies are conducted  for $\ell_\text{gph}$  after all relationships are built. The training procedure can be found in the supplementary material.

\section{Experiments}
\begin{figure}[t]
	\centering
	\includegraphics[width=\linewidth]{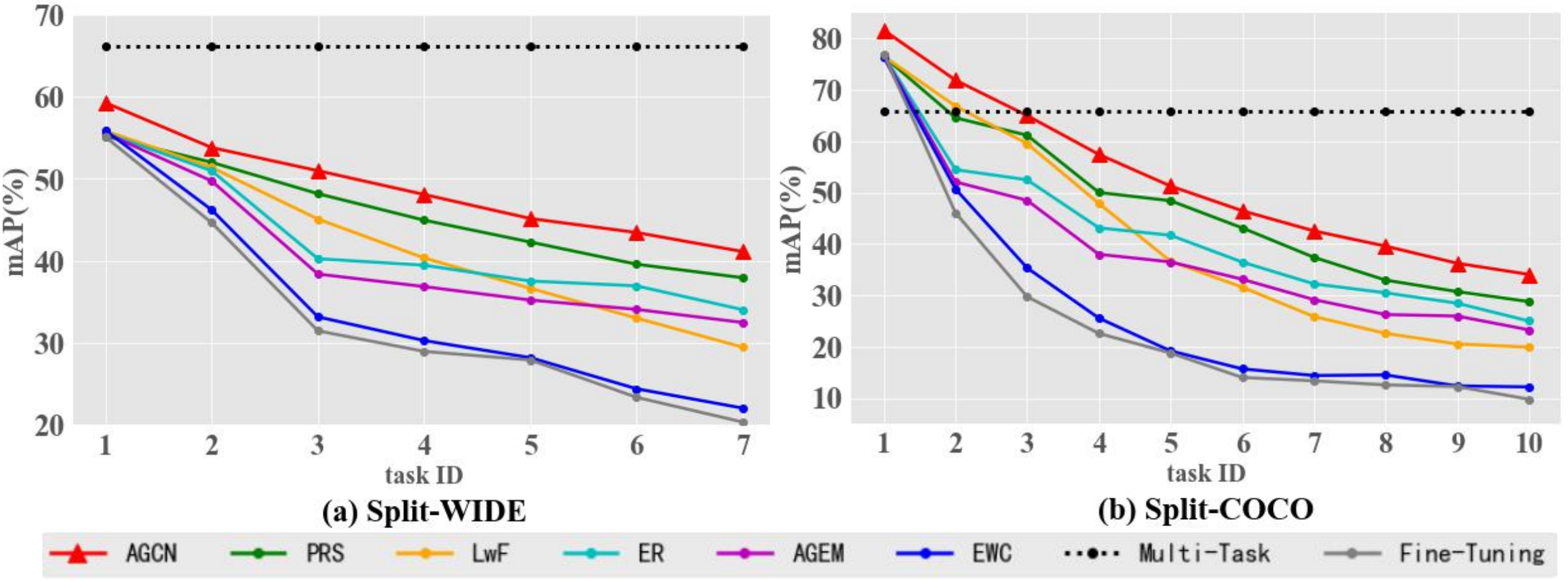}
	\vspace{-20px}
	\caption{mAP (\%) changes on two benchmarks.}
	\label{fig:learning_curves}
\end{figure}

\begin{table}[t]
	\centering
	\caption{We report 3 main metrics (\%) for multi-label classification after the whole data stream is seen once on Split-WIDE and Split-COCO. The final values are the average of the values running with 5 different random seeds.}
	\resizebox{\linewidth}{!}{
	\begin{tabular}{c|rrr|rrr}
		\toprule
		\multirow{2}{*}{\textbf{Method}} & \multicolumn{3}{c|}{\textbf{Split-WIDE}} & \multicolumn{3}{c}{\textbf{Split-COCO}}  \\ 
		\cline{2-7} 
        \cline{2-7} 
		&\textbf{mAP} $\uparrow$&\textbf{CF1} $\uparrow$&\textbf{OF1} $\uparrow$&\textbf{mAP} $\uparrow$&\textbf{CF1} $\uparrow$&\textbf{OF1} $\uparrow$\\
		\hline
		\hline
		\textbf{Multi-Task}&66.17&61.45&71.57
		&65.85&61.79&66.27
		 \\

		\hline
		\textbf{Fine-Tuning}&20.33&19.10&35.72 
		&9.83&10.54&28.83
		\\
		\rowcolor{mygray}
		Forgetting $\downarrow$ &40.85&31.20&15.10
		  &58.04&63.54&20.60
		\\
		 \hline

		\textbf{EWC}~\cite{kirkpatrick2017overcoming}&22.03&22.78&35.70
		&12.20&12.50&29.67
		
 \\
		\rowcolor{mygray}
		Forgetting $\downarrow$&34.86&28.18&15.17
		&45.61&55.44&19.85
		
 \\
		 \hline
		\textbf{LwF}~\cite{li2017learning}&29.46&{29.64}&{42.69}
		&{19.95}&{21.69}&{40.68}
		 \\
		\rowcolor{mygray}
		Forgetting $\downarrow$&20.26&18.99&5.73
		&41.16&39.85&11.43
		 \\
		 \hline
		\textbf{AGEM}~\cite{chaudhry2018efficient}&32.47&33.28&38.93
		&23.31&27.25&37.94
		\\
		\rowcolor{mygray}
		Forgetting $\downarrow$&16.42&15.71&9.73
		&34.52&18.92&12.94
		 \\
		 \hline
		\textbf{ER}~\cite{rolnick2019experience}&{34.03}&{34.94}&{39.37}
		&25.03&30.54&38.38
		 \\
		\rowcolor{mygray}
		Forgetting $\downarrow$&15.15&11.80&8.61
		&33.46&17.28&12.34 
	 \\
		 \hline
		 \textbf{PRS}~\cite{kim2020imbalanced}&{37.93}&{21.12}&{15.64}
		&{28.81}&18.40&13.86 \\
		\rowcolor{mygray}
		Forgetting $\downarrow$&13.59&51.09&62.90
		&30.90&54.36&52.51 \\

		\hline
		\textbf{AGCN (Ours)}&\textbf{41.12}&\textbf{38.27}&\textbf{43.27}
		&\textbf{34.11}&\textbf{35.49}&\textbf{42.37}\\
		\rowcolor{mygray}
		Forgetting $\downarrow$&11.22&5.43&4.28
		 &23.71&14.79&8.16\\
		\bottomrule
	\end{tabular}}
\label{tab:COCO}
\end{table}

\begin{table}[t] 
    \centering
    \caption{Ablation studies (\%) for  ACM $\mb{A}^t$  used to model intra- and inter-task label relationships on Split-COCO.}
    \resizebox{0.8\linewidth}{!}{
        \begin{tabular}{c|cc|ccc}
        \toprule
            
             &$\mb{A}^{t-1}$ \& $\mb{B}^t$  & $\mb{R}^t$ \& $\mb{Q}^t$  &\textbf{mAP} $\uparrow$&\textbf{CF1} $\uparrow$&\textbf{OF1} $\uparrow$  \\
            \hline
           1& $\surd$ & $\times$ &  31.52 & 30.37 & 34.87   \\
           2& $\surd$ & $\surd$ & 34.11 & 35.49 & 42.37  \\
            \bottomrule
    \end{tabular}}
    \label{tab:ablation}
    \vspace{-15px}
\end{table}

\subsection{Dataset construction}
The datasets, namely Split-COCO and Split-WIDE, for LML image recognition, are constructed using two large-scale popular multi-label image datasets, \ie,  MS-COCO~\cite{lin2014microsoft} and NUS-WIDE~\cite{chua2009nus}.\\
\noindent
\textbf{Split-COCO}.
We choose the 40 most frequent concepts from 80 classes of MS-COCO to construct Split-COCO, which has 65082 examples for training and 27,173 examples for validation.
The 40 classes are split into 10 different and non-overlapping tasks, each of which contains 4 classes.

\noindent
\textbf{Split-WIDE}. 
As another multi-label image dataset, NUS-WIDE has a larger scale than MS-COCO.
Following~\cite{jiang2017deep}, we choose the 21 most frequent concepts from 81 classes of NUS-WIDE to construct the Split-WIDE, which has 144,858 examples for training and 41,146 examples for validation.
We split the Split-WIDE into 7 tasks, where each task contains 3 classes. 
More details about the two datasets can be found in the supplementary material.

\subsection{Baseline methods}

We compare our method with several important and state-of-art lifelong learning methods including
(1) \textbf{EWC}~\cite{kirkpatrick2017overcoming}, which regularizes the training loss to avoid catastrophic forgetting;
(2) \textbf{LwF}~\cite{li2017learning}, which uses the distillation loss by saving task-specific parameters;
(3) \textbf{AGEM}~\cite{chaudhry2018efficient} and (4) \textbf{ER}~\cite{rolnick2019experience}, which save a few of training data from the old tasks and retrains them in the current training.
(5) \textbf{PRS}~\cite{kim2020imbalanced}, which uses a different rehearsal strategy to study the imbalanced problem.
Note that PRS studies similar problems with us, but they focus more on the imbalanced problem but ignore the label relationships and the problem of partial labels for LML image recognition.
Following existing approaches like ~\cite{chaudhry2018efficient,kim2020imbalanced}, we use a multi-task baseline, \textbf{Multi-Task}, which is trained on a single pass over shuffled data from all tasks, it can be seen as an upper bound performance. We also compare with the \textbf{Fine-Tuning}, which performs online training without any lifelong learning technique, thus it can be regarded as a lower-bound performance. Note that, to extend some LSL methods to LML, we turn the final softmax layer in each of these methods to a sigmoid. Other details follow their original settings. More implementation details can be found in the supplementary material.

\subsection{Evaluation metrics}

\noindent
\textbf{Multi-label evaluation}. Following the traditional multi-label learning \cite{wang2016cnn,jin2016annotation,chen2019multi}, 3 metrics are used for evaluation in LML image recognition.
(1) the mean average precision (\textbf{mAP}) over all labels;
(2) the per-class F1-measure (\textbf{CF1});
(3) the overall F1-measure (\textbf{OF1}). 
The mAP, CF1 and OF1 are relatively more important for multi-label performance evaluation.
In particular, the three multi-label metrics are computed when all tasks are trained done, \ie, the final score.

\noindent
\textbf{Forgetting measure}~\cite{chaudhry2018riemannian}. 
This score denotes the value difference of the above three multi-label metrics between the final score and the score when the task was first trained done.
For example, the forgetting measure of mAP for the task $t$ can be computed by its performance difference between task $T$ and $t$ was trained.
Note that the negative values of forgetting mean no forgetting and improved performance at the training phase.

            

\begin{table}[t] 
    \centering
    \caption{AGCN ablation studies (\%) for loss weights and relationship-preserving loss on Split-COCO.}
    \resizebox{0.7\linewidth}{!}{
        \begin{tabular}{ccc|ccc}
            \toprule
             $\lambda_1$ & $\lambda_2$   &$\lambda_3$   &\textbf{mAP} $\uparrow$&\textbf{CF1} $\uparrow$&\textbf{OF1} $\uparrow$ \\
            \hline
            \hline
             
             $0.05$ & $0.95$   & $0$    & 29.90 & 31.80 & 37.12  \\
             \rowcolor{mygray}
              & Forgetting $\downarrow$&  & 29.24 & 24.88 & 19.67  \\
              \hline
             $0.07$ & $0.93$   & $0$    & 30.99 & 32.03 & 39.31   \\
             \rowcolor{mygray}
              &  Forgetting  $\downarrow$&    & 28.28 & 22.55 & 13.88   \\
              \hline

             $0.09$ & $0.91$   & $0$    & 29.71 & 32.71 & 38.91 \\
             \rowcolor{mygray}
              &  Forgetting  $\downarrow$&     & 29.97 & 21.79 & 16.49 \\
             \bottomrule
             $0.07$ & $0.93$   & $ 10^4 $    & 33.05 & 33.31 & 41.04   \\
             \rowcolor{mygray}
             &  Forgetting  $\downarrow$&     & 26.41 & 20.99 & 11.38 \\
             \hline
             $0.07$ & $0.93$   & $ 10^5 $    & \textbf{34.11} & \textbf{35.49} & 42.37   \\
             \rowcolor{mygray}
             &  Forgetting  $\downarrow$&     & 23.71 & 14.79 & 8.16 \\
             \hline
             $0.07$ & $0.93$   & $ 10^6 $    & 33.71 & 33.05 & \textbf{42.62}   \\
             \rowcolor{mygray}
             &  Forgetting  $\downarrow$&     & 25.69 & 21.30 & 7.89 \\
            \bottomrule
    \end{tabular}}
    \label{tab:lambda}
\end{table}

\begin{figure}[t]
	\centering
	\includegraphics[width=0.7\linewidth]{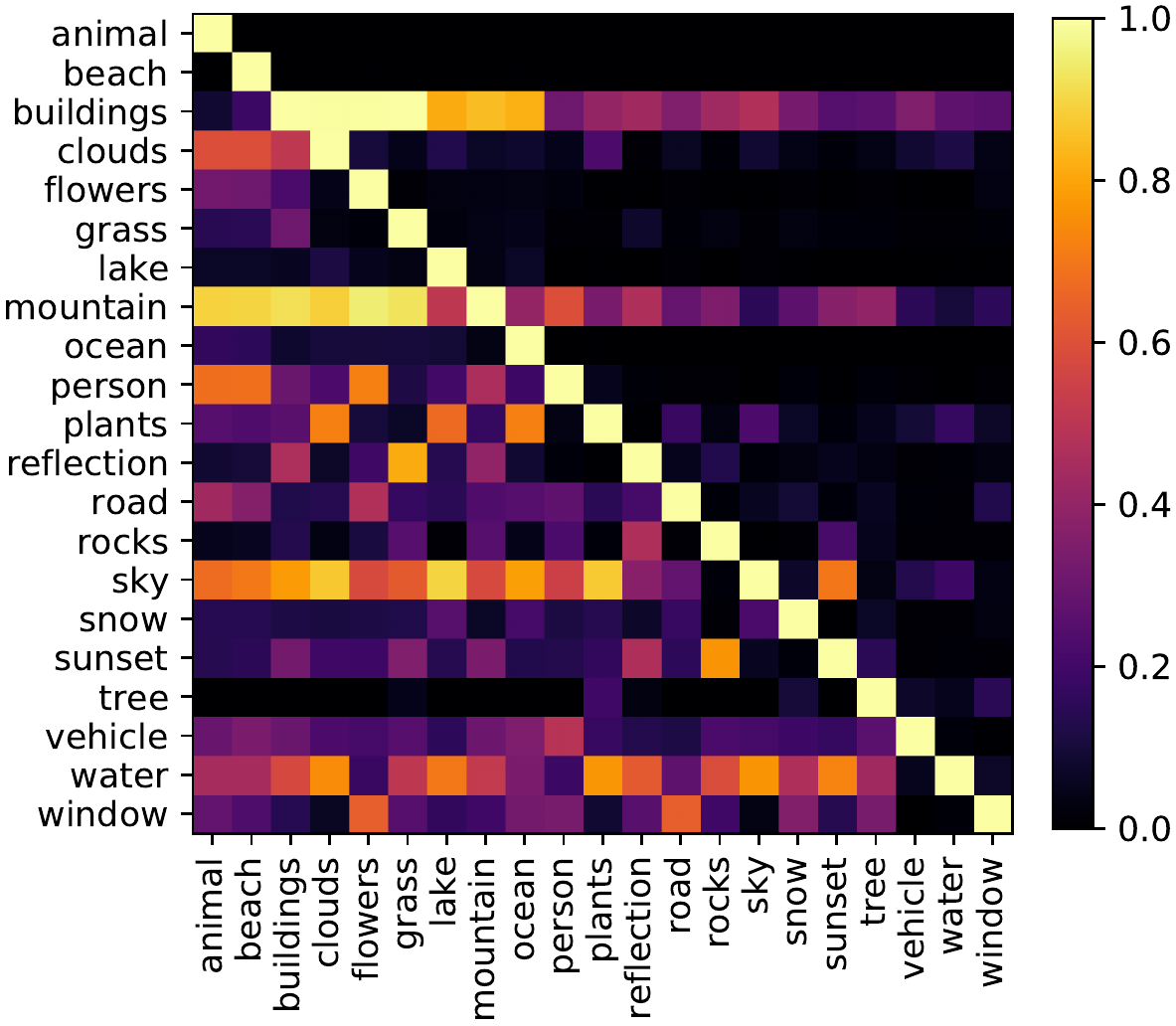}
	\caption{ACM visualization on Split-WIDE.}
	\label{fig:visualization_NUS}
	\vspace{-15px}
\end{figure}

\subsection{Main results}


In Tab.~\ref{tab:COCO}, our method shows better performance than the other state-of-art methods on the three metrics as well as the forgetting value evaluated after task $T$. 
On Split-COCO, the AGCN outperforms the best of the state-of-art method PRS by a large margin (34.11\% vs. 28.81\%). 
The AGCN shows better performance than the others on Split-WIDE (41.12\% vs. 37.93\%), which suggests that AGCN is also effective on the large-scale multi-label dataset.
As shown in Fig.~\ref{fig:learning_curves}, which illustrates the mAP changes as tasks are being learned on two benchmarks. 
Easy to find, the proposed AGCN is better than other state-of-art methods through the whole LML process.

\subsection{ACM Visualization}
The ACM visualization on Split-WIDE is shown in Fig.~\ref{fig:visualization_NUS}.
The dependency between two classes with higher correlation has larger weights than irrelevant ones, which means the intra- and inter-task relationships can be well constructed even the old classes are unavailable. 
For example, the label ``sky'' is closely related to ``clouds'', ``sunset'' and ``lake''; the label ``person'' is closely related to ``beach'' and ``flowers''. 
We can also find ``person'' and ``vehicle'' should be closely related but not, which we think is because of the low co-occurrence frequency of the two classes in the dataset.


\subsection{Ablation studies}
\textbf{ACM effectiveness}. 
In Tab.~\ref{tab:ablation}, if we do not build the relationships cross old and new tasks, the performance of AGCN (the first row) is already better than other non-AGCN methods, for example, 31.52\% vs. 28.81\% in mAP.
This means only intra-task label relationships are effective for LML image recognition. 
When the inter-task block matrices $\mb{R}^t$ and $\mb{Q}^t$ are available, AGCN with both intra- and inter-task relationships (the second row) can perform even better in all three metrics, which means the inter-task relationships established by soft label statistics can further enhance the multi-label recognition. 

\noindent
\textbf{Hyperparameter selection}.
Then, we analyze the influences of loss weights and relationship-preserving loss on Split-COCO as shown in Tab.~\ref{tab:lambda}. 
When $\lambda_1=0.07$, $\lambda_2=0.93$, the performance is better than others. 
By adding the relationship-preserving loss $\ell_\text{gph}$, the performance obtains larger gains, which means the mitigation of catastrophic forgetting of relationships is quite important for LML image recognition. 
We select the best $\lambda_3$ as the hyper-parameters, \ie, $\lambda_3=10^5$ for LML image recognition.

\section{Conclusion}
LML image recognition is a new paradigm of lifelong learning, in this paper, a novel AGCN based on an auto-updated expert mechanism is proposed to solve the problems in LML image recognition. The key challenges are to construct label relationships and reduce catastrophic forgetting to improve overall performance.
We construct the label relationships by solving the partial label problem with soft labels generated by the expert network. 
We also mitigate the relationship forgetting by the proposed relationship-preserving loss. 
In this way, the proposed AGCN can connect previous and current tasks on all seen classes in LML image recognition.
Extensive experiments demonstrate that AGCN can capture well the label dependencies and effectively mitigate the catastrophic forgetting thus achieving better recognition performance. 

\section{Acknowledgment}
This work was supported by the Natural Science Foundation of China (No. 61876121) and Postgraduate Research \& Practice Innovation Program of Jiangsu Province (No. 092192701). The authors would like to thank constructive and valuable suggestions for this paper from the experienced reviewers and AE.

\bibliographystyle{IEEEbib}
\bibliography{icme2021template}

\newpage

\section*{Supplementary Material}}

\section*{A.1 The whole algorithm}
An algorithm for LML is presented in Alg.(\ref{alg:main}) to show the detailed training procedure of AGCN.
Given the training dataset $\mc{D}_\text{trn}^t$ and the initialed graph node $\mb{H}^{t,0}$:
(1) The intra-task correlation matrix $\mb{A}^1$ is constructed by the statistics of hard labels $\mb{y}$, then based on the $\mb{A}^1$, and the prediction $\mb{\hat{y}}$ is generated by the $\text{AGCN}$ model and the $\text{CNN}$ model.
(2) When $t>1$, the ACM $\mb{A}^t$ is augmented via soft labels $\mb{\hat{z}}$ and the Bayes' rule. Based on the $\mb{A}^t$, $\text{AGCN}$ model can capture intra- and inter-task label dependencies. Then, $\mb{\hat{z}}$  and $\mb{G}^{t-1}$ as target features to build $\ell_\text{dst}$ and $\ell_\text{gph}$ respectively.
(3) The $\text{CNN}_\text{xpt}$ and $\text{AGCN}_\text{xpt}$ are updated by the trained $\text{CNN}$ and $\text{AGCN}$ respectively. After all the data has been trained once, the final model $\text{AGCN}$ and $\text{CNN}$ are returned.
The training and testing label sets of task $t$ are shown in Tab.~\ref{tab:scenarios}.

\begin{algorithm}[ht]
	\caption{Training procedure of AGCN.}
	\label{alg:main}
	\begin{algorithmic}
		\REQUIRE $\mc{D}^t_\text{trn}, \{\mb{H}^{t,0}\}_{t\in[1,T]}$ \\
		
		\FOR{$t=1$ {\bfseries to} $T$}
		\FOR{$(\mb{x},\mb{y})\sim\mc{D}^t_\text{trn}$}
		\IF{$t=1$}
		\STATE Compute $\mb{A}^1$ with $\mb{y}$ using Eq.(5)\;
		\STATE $\mb{\hat{y}} = \sigma\left({\text{AGCN}(\mb{A}^1, \mb{H}^{1,0})}\otimes \text{CNN}\left(\mb{x}\right)\right)$\;
		\STATE $\ell=\ell_\text{cls}(\mb{y},\mb{\hat{y}})$\;
		\ELSE
		\STATE $\hat{\mb{z}} = \sigma\left(\text{AGCN}_\text{xpt}(\mb{A}^{t-1}, \mb{H}^{t-1,0}) \otimes \text{CNN}_\text{xpt}\left(\mb{x}\right)\right)$\;
		\STATE	Compute $\mb{B}^t$ with $\mb{y}$ using Eq.(5)\;
		\STATE	Compute $\mb{R}^t$ and $\mb{Q}^t$ using Eq.(6) and Eq.(7)\;				
		\STATE ${\mb{A}}^{t}=\begin{bmatrix} {\mb{A}}^{t-1} & \mb{R}^t \\ \mb{Q}^t & \mb{B}^t \end{bmatrix}$ \\
		\STATE $\mb{H}^t=\text{AGCN}(\mb{A}^t,\mb{H}^{t,0})$ 				\STATE ${\mb{G}^{t-1}}=\text{AGCN}_\text{xpt}(\mb{A}^{t-1},\mb{H}^{t-1,0})$\;		 
		\STATE $\mb{\hat{y}} = \sigma\left({\text{AGCN}(\mb{A}^t, \mb{H}^{t})}\otimes \text{CNN}\left(\mb{x}\right)\right)$\;
		\STATE	$\ell=\lambda_1\ell_\text{cls}(\mb{y},\mb{\hat{y}}_\text{new})+\lambda_2\ell_\text{dst}(\mb{\hat{z}},\mb{\hat{y}}_\text{old})$\STATE$~~~~~~+\lambda_3\ell_\text{gph}({\mb{G}}^{t-1},\mb{H}^{t})$\;
		\ENDIF
		\STATE Update $\text{AGCN}$ and $\text{CNN}$ by minimizing $\ell$ \\
		\ENDFOR
		\STATE $\text{CNN}_\text{xpt}=\text{CNN}, \text{AGCN}_\text{xpt}=\text{AGCN}$ \\
		\ENDFOR
		\STATE {\bfseries Return} $\text{AGCN}$, $\text{CNN}$
	\end{algorithmic}
\end{algorithm}

\section*{A.2 Dataset construction}

\begin{table}[t]
	\centering
	\caption{
		Training and testing label sets of task $t$.
	}
	\resizebox{0.6\linewidth}{!}{
		\begin{tabular}{c|c}
			\toprule
			Training     & $\mc{Y}\subseteq\mc{C}^t$  \\ 
			Testing     & $\mc{Y}\subseteq\mc{C}_\text{seen}^t=\mc{C}_\text{seen}^{t-1}\cup\mc{C}^t$   \\ 
			\bottomrule
	\end{tabular}}
	\label{tab:scenarios}
\end{table}

PRS~\cite{kim2020imbalanced} needs more low-frequency classes to study the imbalanced problem, they curate 4 tasks with 70 classes for lifelong multi-label learning.
Multi-labelled datasets inherently have intersecting concepts among the data points. Hence, a naive splitting algorithm may lead to a dangerous amount of data loss. This motivates our first objective to minimize the data loss during the split. Additionally, in order for us to test diverse research environments, the second objective is to optionally keep the size of the splits balanced.  
To split the well-known MS-COCO and NUS-WIDE into several different tasks fairly and uniformly, we introduce two kinds of images in the datasets:\\ 
\textbf{Special-labeling}. If an image only has the labels that belong to the task-special class set $\mc{C}^t$ of task $t$, we regard it as a special-labelling image for task $t$.\\ 
\textbf{Mixed-labeling}. If an image not only has the task-specific labels but also has the old labels belong to the class set $\mc{C}^{t-1}_{\text{seen}}$, we regard it as a mixed-labelling image.\\
In LML, because the model just learns the task-specific labels $\mc{C}^t$, the training data is labelled without old labels, so LML will suffer from the label missing problem, which mainly appears in the mixed-labelling image. 
For each task,  a randomly data-splitting approach may lead to the imbalance of special-labelling and mixed-labelling images.
To ensure a proper proportion of special-labelling images and mixed-labelling images, we split two datasets into sequential tasks with the following strategies:
We first count the number of labels for each image.
Then, we give priority to leaving special-label images for each task, the mixed-labelling images are then allocated to other tasks. 
In addition, the larger the total number of labels in all tasks, the lower the proportion of special-labelling images in each task, and the higher the proportion of mixed-labelling images, the more obvious the partial label problem. The dataset construction is presented in Tab.~\ref{tab:dataset}.


\begin{table}[t]
	\centering
	\caption{Dataset construction.}
	\resizebox{.8\linewidth}{!}{
		\begin{tabular}{c|c|c|c}
			\hline
			\multicolumn{1}{c|}{}& \multicolumn{1}{| c|}{}& \multicolumn{1}{|c|}{\textbf{special-label}}& \multicolumn{1}{|c}{\textbf{mixed-label}}\\ 
			\hline
			\textbf{Dataset}&\textbf{Task ID}&\textbf{number}&\textbf{number} \\
			\hline
			\hline
			&1&8511&3101\\
			&2&2772&3528\\
			&3&2720&3508\\
			&4&799&5320\\
			&5&150&6111\\
			Split-COCO&6&603&5500\\
			&7&901&5234\\
			&8&1001&5161\\
			&9&1198&4803\\
			&10&1947&2214\\
			&sum&20602&44480\\
			\hline
			\multirow{8}{*}{Split-WIDE}
			&1&22927&10067\\
			&2&3627&17012\\
			&3&7727&12996\\
			&4&1154&7095\\
			&5&14599&10111\\
			&6&4936&18995\\
			&7&4546&9066\\
			&sum&59516&85342\\
			\hline
	\end{tabular}}
	\label{tab:dataset}
\end{table}

\section*{A.3 Augmented Graph Convolutional Network}
Based on the ACM, we use the AGCN to capture the label dependencies across tasks.
Each node will be initialized by the Glove embedding and updated layer-by-layer:

\begin{equation}\label{eq:gcn}
		\begin{aligned}
		\mb{H}^{t,l+1}&=\text{AGCN}\left(\mb{A}^t,\mb{H}^{t,l}\right) =h\left(\mb{A}^t\mb{H}^{t,l}{\mb{W}}^{t,l}\right)\\
		&=h\left(
		\begin{bmatrix} \mb{A}^{t-1} & \mb{R}^t \\ \mb{Q}^t & \mb{B}^t \end{bmatrix}
		\mb{H}^{t,l}{\mb{W}}^{t,l}
		\right),
	\end{aligned}
\end{equation}
where AGCN is a two-layer stacked GCN model, and we set $l\in\{0,1\}$ in our study following the method in~\cite{chen2019multi}, $h(\cdot)$ denotes a non-linear operation, $\mb{W}^{t,l}$ is a transformation matrix to be learned. For convenience, the function that the two-layer AGCN model wants to learn in task $t$ is denoted as $\text{AGCN}(\cdot,\cdot)$ with parameters $\mb{W}$. The output of AGCN is denoted as $\mb{H}^t = \mb{H}^{t,2}$, $\mb{W}^{t,1}\in\mbb{R}^{d \times d'}$, $\mb{W}^{t,2}\in\mbb{R}^{d' \times D}$, $d$ denotes the initial embedding dimension, and $D$ represents the image feature dimension. Following~\cite{chen2019multi}, $d'=1024$ and $D=2048$ in our method.

\section*{A.4 Implementation details} 
Following existing multi-label image classification methods, We employ ResNet101 as the image feature extractor pre-trained on ImageNet. We adopt Adam as the optimizer of network with $\beta_1=0.9$, $\beta_2=0.999$, and $\epsilon=10^{-4}$. Our AGCN consists of two GCN layers with output dimensionality of 1024 and 2048, respectively. The input images are random cropped and resized into $448\times448$ with random horizontal flips for data augmentation. The network is trained for a single epoch. 

\section*{A.5 Evaluation metrics}

\noindent
\textbf{Multi-label Evaluation.}
Following the traditional multi-label learning, we compute the overall and per-class precision, recall and F1 score as:
\begin{equation}
	\label{eq:metrics}
	\begin{aligned}
		& OP=\frac{\sum_iN_i^c}{\sum_iN_i^p},~~~~~~~ 
		~~~~~~~~CP=\frac{1}{C}\sum_i\frac{N_i^c}{N_i^p},\\
		& OR=\frac{\sum_iN_i^c}{\sum_iN_i^g},~~~~~~~ 
		~~~~~~~~CR=\frac{1}{C}\sum_i\frac{N_i^c}{N_i^g},\\
		& OF1=\frac{2 \times OP \times OR}{OP+OR},~ 
		CF1=\frac{2 \times CP \times CR}{CP+CR},	
	\end{aligned}
\end{equation}
where $i$ is the class label and $C$ is the number of labels. $N_c^i$ is the number of correctly predicted images for class $i$, $N^p_i$ is the number of predicted images for class $i$ and $N_i^g$ is the number of ground-truth for class $i$. 
The mAP is computed by the average CP across all data.

\noindent

\noindent
\textbf{Forgetting Measure}.~($F_t\in[-1,1]$) 
Average forgetting after the model has been trained continually  up till task $t\in\{1,\cdots,T\}$ is defined as:
\begin{equation} \label{eq:forget}
	F_{t}=\frac{1}{t-1} \sum_{j=1}^{t-1} f_{j}^{t},
\end{equation}
where $f_{j}^{t}$ is the forgetting on task $j$ after the model is trained up till task $t$ and computed as
\begin{equation}
	f_{j}^{t}=\max _{l \in\{1, \cdots, k-1\}} a_{l, j}-a_{t, j},
\end{equation}
where $a$ denotes every metric in LML like mAP, CF1 and OF1.

\section*{A.6 Visualization of ACM evolution}
\begin{figure}[h]
	\centering
	\includegraphics[width=\linewidth]{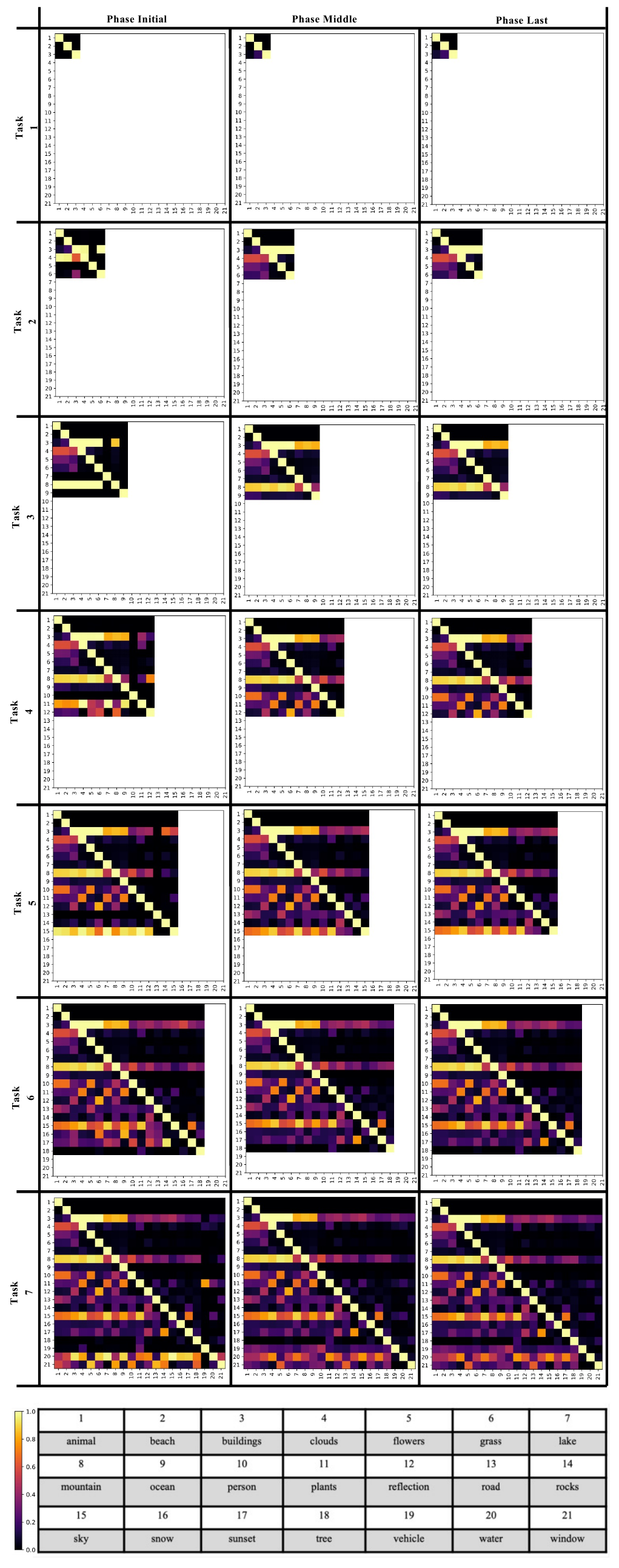}
	\caption{ACM augmented process visualization of Split-WIDE. It shows the intra- and inter-task label relationships are constructed completely and preserved well.}
	\label{fig:m_COCO}
\end{figure}

As shown in Fig.~\ref{fig:m_COCO}, we show the ACM visualizations on Split-WIDE for LML.  
Several observations can be obtained: 
1) the ACM will be augmented from $\mb{A}^1$ to $\mb{A}^7$ along with the task sequences and the intra- and inter-task relationships can be obtained; 
2) the dependency between two classes with higher correlation (higher co-occurrence frequency) has larger weights else smaller ones; 
3) the intra- and inter-task relationships can also be constructed even the old classes are unavailable; 
and 4) form $\mb{A}^1$ to $\mb{A}^7$, the label relationships can be preserved well. For example, the label "sky" is closely related to "clouds", "sunset" and "lake"; the label "person" is closely related to "beach" and "flowers".

\end{document}